\documentclass[10pt,twocolumn,letterpaper]{article}

\usepackage{times}
\usepackage{graphicx}
\usepackage{amsmath}
\usepackage{amsthm}
\usepackage{amssymb}
\usepackage{epsfig}

\usepackage{helvet}  %Required
\usepackage{courier}  %Required
\usepackage{url}  %Required
\usepackage{color}
\usepackage{subfigure} 
\usepackage{algorithm}
\usepackage{algorithmic}

\usepackage{multibib}

\usepackage{comment}
\usepackage{epstopdf}
\usepackage{array}
\usepackage{wrapfig}
\usepackage{adjustbox}
\usepackage{capt-of}
\usepackage{float}
\usepackage{rotating}
\usepackage{multirow}
\usepackage{mathtools}
\usepackage{bm}
\renewcommand{\vec}[1]{\mbox{\bm{$#1$}}}
\usepackage[pagebackref=true,breaklinks=true,colorlinks,bookmarks=false]{hyperref}
\newcommand{\mat}[1]{{\pmb #1}}
\newcommand{\argmax}{\operatornamewithlimits{argmax}}
\newcommand{\argmin}{\operatornamewithlimits{argmin}}

\newtheorem{defin}{Definition}

\usepackage{cvpr}

\def\cvprPaperID{0} % *** Enter the CVPR Paper ID here

\cvprfinalcopy % *** Uncomment this line for the final submission

\ifcvprfinal\fi

\begin{document}

%%%%%%%%% TITLE
%\title{Adversarially Optimizing Intersection over Union for Object Localization Tasks}
%\title{Adversarial Data Augmentation for Object Localization}
%\title{Optimal Data Augmentation for Object Localization}
\title{ADA: A Game-Theoretic Perspective on Data Augmentation for Object Detection}
\author{Sima Behpour\\
University of Illinois at Chicago\\
{\tt\small sbehpo2@uic.edu}
\and
Kris M. Kitani\\
Carnegie Mellon University\\
{\tt\small kkitani@cs.cmu.edu }
\and
Brian D. Ziebart\\
University of Illinois at Chicago\\
{\tt\small bziebart@uic.edu }
}

\maketitle

\begin{abstract}
The use of random perturbations of ground truth data, such as random translation or scaling of bounding boxes, is a common heuristic used for data augmentation that has been shown to prevent overfitting and improve generalization. Since the design of data augmentation is largely guided by reported best practices, it is difficult to understand if those design choices are optimal. To provide a more principled perspective, we develop a game-theoretic interpretation of data augmentation in the context of object detection. We aim to find an optimal adversarial perturbations of the ground truth data (i.e., the worst case perturbations) that forces the object bounding box predictor to learn from the hardest distribution of perturbed examples for better test-time performance. We establish that the game theoretic solution, the Nash equilibrium, provides both an optimal predictor and optimal data augmentation distribution. We show that our adversarial method of training a predictor can significantly improve test time performance for the task of object detection. On the ImageNet object detection task, our adversarial approach improves performance by over 16\% compared to the best performing data augmentation method.\vspace{-3mm}
\end{abstract}

\vspace{-3mm}
\section{Introduction}

\begin{figure}
\centering
\includegraphics[width= 0.9\columnwidth]{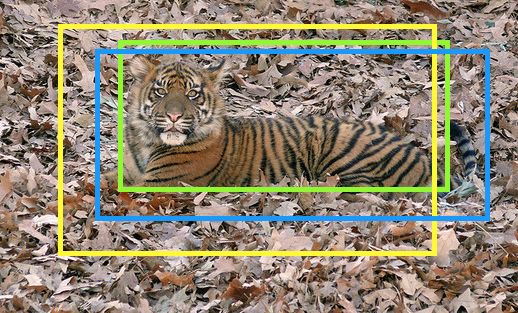}
\caption{Tiger localization example with three different bounding box annotations illustrates ambiguity in `ground truth' labels.}
\label{fig:teaser}
\vspace{-5mm}
\end{figure}

There is no guarantee that human-labeled `ground-truth' annotations of an image dataset are error free. Consider the bounding box annotations of three annotators of the image in Figure \ref{fig:teaser}. Do all boxes contain the object? Are all three bounding boxes equally correct? Is there one bounding box which is most helpful for learning a detection model? These questions highlight the ambiguity in the annotation task. In response, many helpful heuristics have been utilized in the literature to obtain more consistent annotations. To deal with inter-annotator disagreement \cite{welinder2010multidimensional,nowak2010reliable,szegedy2016inception}, previous work has relied primarily on reasonable heuristics for augmenting the ground truth through consensus \cite{upchurch2016interactive,saragih2011deformable,zhu2012face,Sagonas_2013_CVPR_Workshops}. Despite these efforts, it is not clear if there is a principled approach for identifying the optimal ground truth distribution in the context of supervised learning. 

To address in part the uncertainty of the `ground truth' annotations, dataset augmentation methods can be used to synthesize new annotations of images by perturbing annotations. In fact, heuristic data augmentation preprocessing such as random translation, flipping or scaling, has been shown to be essential for many modern visual learning tasks using deep networks. However, manually choosing perturbations to improve performance can be an error-prone process. While increasing the modes of data perturbations may effectively increase the amount of training data, it can also cause the learned predictor to drift. In this way, current techniques for data augmentation are more of an art than a science.

Towards a more principled approach to data augmentation, we propose to integrate annotation perturbations directly into the learning process. We do this by introducing an adversarial function that generates maximally perturbed version of the ground truth, which makes it as hard as possible for the predictor to learn. The adversary, however, is not completely free to perturb the data. It must retain certain feature statistics (\emph{e.g.}, the features of the new bounding box distribution should still be close to the features of the original bounding box). Formally, we pose the data augmentation problem as a zero-sum game between a player (the predictor model) seeking to maximize performance and a constrained adversary (augmented data distribution) seeking to minimize expected performance \cite{grunwald2004game,asif2015adversarial,wang2015adversarial}.

To the best of our knowledge, this is the first work to provide a theoretic basis for data augmentation in terms of an adversarial two player zero-sum game. As a consequence of our game-theoretic formulation, we develop a novel adversarial loss function that identifies the optimal data augmentation strategy which leads to the most robust predictor possible (\emph{i.e.}, trained for the worst case perturbation of data). In our experiments, we focus on the task of object detection and show that our proposed adversarial data augmentation technique consistently improves performance over various competing loss functions, data augmentation levels, and deep network architectures.

\section{Related Work}

It is common to assume that the ground truth is singular and error-free.
However, disagreement between annotators is a widely-known problem for many computer vision tasks \cite{welinder2010multidimensional}, as well as a major concern \cite{nowak2010reliable} when constructing an annotated computer vision corpora.
In large part, the difficulty arises because the set of possible ``ground truth'' annotations is typically extremely large for vision tasks.  It is a powerset of possible descriptions (\emph{e.g.}, words, noun phrases) in annotation tasks, multi-partitions of the pixels (exponential in the number of pixels) in segmentation tasks, and the possible bounding boxes (quadratic in the number of pixels) for localization tasks.

Methods to form a ``consensus'' annotation and to improve the annotation process through crowd-sourcing have been developed by averaging or combining together different independent annotations \cite{upchurch2016interactive}, verifying annotations with other independent annotators \cite{saragih2011deformable}, and other strategies \cite{zhu2012face,Sagonas_2013_CVPR_Workshops}.  For example, the ILSVRC2012 image dataset employs boundary box drawing, quality verification, and coverage verification as three separate subtasks \cite{su2012crowdsourcing} in a crowd-sourcing pipeline. In the construction of that dataset, proposed bounding boxes are rejected 37.8\% of the time \cite{su2012crowdsourcing}, illustrating the inherent disagreement between annotators and the uncertainty of the task. Despite the added safeguards of the verification process, recent evaluations have also been performed
by removing a substantial fraction of the training examples that are considered to have poor quality bounding boxes
\cite{szegedy2016inception,gokberk2014multi,xiao2015learning,sukhbaatar2014learning,reed2014training}.

Many state-of-the-art methods for object detection  \cite{krizhevsky2012imagenet} are based on CNN, and incorporate other improvements such as the use of very large scale datasets, more efficient GPU computation, and data augmentation \cite{chatfield2014return}. Recently, most of the literature on data augmentation studies effective data augmentation methods for CNN features that increases the performance of different tasks (\emph{e.g.}, classification, object recognition) \cite{taylor2017improving,paulin2014transformation,masi2016we,he2015spatial}. Chatfield \cite{chatfield2014return} applies the data augmentation techniques commonly applied to CNN-based methods to shallow methods and shows an analogous performance boost \cite{chatfield2014return}. Paulin et al. \cite{paulin2014transformation} claim that given a large set of possible transformations, all transformations are not equally informative and adding uninformative transformations increases training time with no gain in accuracy. They propose Image Transformation Pursuit (ITP) algorithm for the automatic selection of a compact set of transformations.

Complementary to our work, data augmentation can also be used to guard against adversarial attacks \cite{goodfellow2014generative,moosavi2016universal,guo2017countering,joon2017adversarial,tramer2017ensemble}. Total variance minimization and image quilting are presented as very effective defenses adversarial-example attacks
on image-classification systems \cite{guo2017countering}. The strength of these data augmentation lies in their non-differentiable nature and their inherent randomness resulting difficult defenses for an adversary. Our work is different in that we seek to optimize the data augmentation process as part of a supervised learning problem.

%=================================================================================%

\section{Problem Formulation}\label{sec:formulation}

In order to understand the underlying theory of adversarial data augmentation proposed in this work, we must first understand the role of the \textit{annotation distribution}, $p(y|\vec{x})$, which describes the distribution over labels $y$ (\emph{e.g.}, a bounding box annotation) given a feature vector $\vec{x}$ (\emph{e.g.}, an RGB image). Note that a training dataset $\mathcal{D} = \{y_n,\vec{x}_n\}_{n=1}^{N}$, induces an annotation distribution $p(y|\vec{x})$. In other words, each label $y_n$ in the training set can be interpreted to be a sample from the annotation distribution which is conditioned on a feature vector $\vec{x}\in\mathbb{R}^{D}$. When there is absolute certainty in the ground truth annotation, the annotation distribution $p(y|\vec{x})$ is an indicator function where it is one for the true label $y^*$ and zero otherwise.

\subsection{Data Augmentation}

The process of data augmentation is a method of altering the annotation distribution. A typical method for data augmentation generates new examples $\tilde{\mathcal{D}}$ by perturbing the training data $\mathcal{D}$. For example, if the label is a structured output like a bounding box (\emph{i.e.}, a vector of four values), we can generate a new structured label $\tilde{y}$ for the same image by slightly perturbing the original `ground truth' bounding box $y^*$. This data augmentation process creates a new underlying annotation distribution $\tilde{p}( y | \vec{x})$. Since data augmentation can be used to generate multiple new labels for the same feature vector $\vec{x}$, the annotation probability $p(y|\vec{x})$ becomes a soft distribution over labels.

Now if we are given a loss function $\ell(\hat{y}, y)$ describing the distance between an estimated label $\hat{y}$ and annotation label $y$, we can compute the expected loss of the estimated label under the annotation distribution as:
\begin{align}
    \sum_{y \in \mathcal{Y}} P(y|\vec{x})
\ell(\hat{y}, y).
\end{align}
Notice that the expected loss is the smallest when the estimated label matches the annotation distribution. Conversely, the expected loss grows larger when the estimated label is far from the annotation distribution. It is important to note here that this marginalization over the annotation distribution is rarely made explicit in the loss function in most modern supervised learning objective functions because the distribution is assumed to be an indicator function at the `ground truth' label.

Now consider the probabilistic predictor $f(y|\vec{x})$ which maps a feature vector $\vec{x}$ to a distribution over labels $y$. The expected loss over the entire dataset $\mathcal{D}$ under the predictor distribution and annotation distribution is defined as:
\begin{align}
    \min_{f \in \Gamma} \sum_{x \in \mathcal{D}}\overbrace{\sum_{y'}  f(y'|\vec{x}) \sum_{y}P(y|\vec{x}) 
    \ell(y', y)}^{\text{expected loss for input }\vec{x}}. \label{eq:expectedLoss}
\end{align}
The goal of supervised learning is to find the optimal predictor $f$ (from some set of predictors $\Gamma$), that minimizes the above expected loss over the labeled training data. Understanding this verbose form of the supervised learning objective function is critical for the formulation that follows.

\subsection{Adversarial Data Augmentation}

\begin{figure}[tb]
\centering
\includegraphics[width= 1.0\columnwidth]{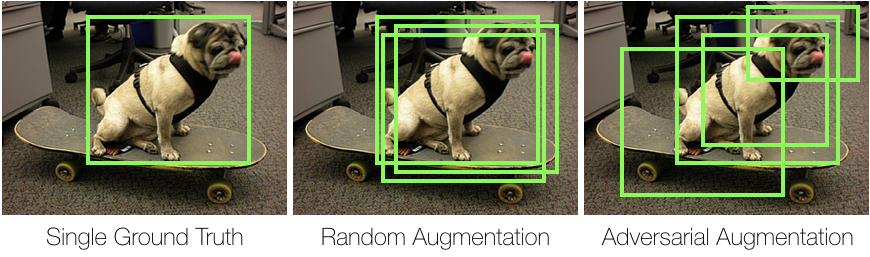}
\caption{Types of annotation distributions. Adversarial augmentation selects bounding boxes that are maximally different from the ground truth but still contain important object features. }
\label{fig:annotation}
\end{figure}

If we adopt a pessimistic view of the annotated data and assume uncertainty in the `ground truth' annotations, we can use data augmentation to perturb the `ground truth' annotations to reflect this uncertainty. We can go further and assume the worst case: that the quality of the annotation distribution is \emph{maximally} perturbed. In other words, we make a strong pessimistic assumption that the annotation distribution was generated by an adversary. By making this worst case assumption, we hypothesize that we can train a more robust predictor that is resilient to large perturbations it might encounter at test time. Figure \ref{fig:annotation} illustrates the three possible choices of annotation distributions for a single image.

More formally, instead of the common Empirical Risk Minimization (ERM) objective of Eq. \eqref{eq:expectedLoss}, we aim to learn a predictor $f$ that optimizes the following adversarial objective function:
\begin{align}
    \min_{f \in \Gamma} \sum_{x\in \mathcal{D}} \sum_{y'}  
    f(y'|\vec{x}) 
    \max_{P(y|x)} \sum_{y} P(y|\vec{x}) 
    \ell(y', y).\label{eq:minimax}
\end{align}
Notice that the maximization sub-problem has been inserted into the objective function which reflects our assumption that the annotation distribution is adversarial (\emph{i.e.}, the worst case distribution). One might quickly notice that this is an unreasonable objective function without some additional constraints because the adversarial annotation distribution can be arbitrarily bad. In the next section we will incorporate constraints that limit the adversary from deviating very far from the original ground truth annotations.

\subsection{Game Formulation}

Our claim is that data augmentation should be included in the learning problem instead of being an independent data pre-processing step. By incorporating data augmentation into the predictor learning problem, we obtain a saddle point optimization problem where we pit a predictor trying to minimize the loss, against an adversarial annotation distribution that is trying to maximize the loss. In this form, the optimization can be seen as a \textit{minimax} problem over a zero-sum two-player game. 

In the language of game theory, the player (predictor) selects a label from a mixed strategy $y' \sim f(y'|\vec{x})$ to minimize the loss, while the opponent (annotation distribution) selects an annotation from the adversarial distribution $y \sim P(y|\vec{x})$ to maximize the loss. The equilibrium point of the game yields both the optimal predictor and an optimal data annotation distribution. The game is zero-sum because the negative loss of the player (predictor) is exactly the gain of the adversary (annotation distribution).

The value or payoff of the game for a particular feature vector $\vec{x}$ is the expected loss of the predictor distribution against the adversary's annotation distribution:
\begin{align}
    \mathbb{E}_{\!\!\!\tiny 
    \begin{array}{c}
    y'|\vec{x}\sim\!\! f\\
    y|\vec{x}\sim\!\! P
    \end{array}}\!\!\!
    \left[ \ell(y',y)\right] \!
    &= 
    \sum_{y',y} 
    f(y'|\vec{x})   \ell(y',y)   P(y|\vec{x}) \\
    &=\mathbf{f}^{\top} \mathbf{G} \mathbf{p}.\label{eq:matrix}
\end{align}
The expected loss of the game can also be written in matrix form, where $\mathbf{f}$ is the vector of probabilities obtained from the predictor over all labels, $\mathbf{G}$ is the game matrix where each element contains the loss between two labels, and $\mathbf{p}$ is the annotation distribution vector.

The adversarial objective function, Eq. (\ref{eq:minimax}) in its current form is problematic because the adversarial annotation distribution is free to perturb the ground truth annotations in arbitrary ways that have no similarity to the original annotations. This can be prevented by constraining the adversarial annotation distribution to choose label distributions in a way that retains feature statistics of the original ground truth annotation. For example, we may want the mean of a set of augmented bounding box annotations to be the same as the mean of the original bounding box annotation. Formally, we can define the first-order statistic of the ground truth data as:
\begin{align}
\mathbb{E}_{y, x \sim \mathcal{D}}
\left[\phi(y, x)\right] 
= \frac{1}{N}\sum_{n=1}^N \phi(y_n, \vec{x}_n),
\end{align}
where $(y_n, \vec{x}_n)$ is the $n^{\text{th}}$ training example in $\mathcal{D}$. We are now ready to define the constrained adversarial optimization problem.

%==============================================%
\begin{defin}  \label{def:game}
The {\bf Primal Adversarial Data Augmentation} (ADA-P) game is: 
\begin{align}
&
\min_{f} \max_{P} 
\mathbb{E}_{
\arraycolsep=0pt\def\arraystretch{1.0}
\tiny\begin{array}{l} 
\vec{x} \sim \mathcal{D},\\
y' |\vec{x} \sim f,\\
y|\vec{x} \sim P
\end{array}}
\left[\ell(y',y) \right]  \text{such that: }\\
&
\mathbb{E}_{
\arraycolsep=0pt\def\arraystretch{1.0}
\tiny\begin{array}{l} 
\vec{x} \sim \mathcal{D},\\
y |\vec{x} \sim P
\end{array}}\left[\phi(y, \vec{x})\right] 
=
\mathbb{E}_{y,x \sim \mathcal{D}}
\left[\phi(y, \vec{x})\right] \notag
\end{align}
where $f(y'|\vec{x})$ and $P(y |\vec{x})$ are distributions over all potential predicted labels for each feature vector $\vec{x}$.
\end{defin}

Due to strong Lagrangian duality \cite{boyd2004convex}, a dual problem with an equivalent solution can be formulated by including the constraint in the objective function using a vector of Lagrangian multipliers, $\theta$. This resulting Lagrangian potential $\theta^{\top} \phi(\cdot,\cdot)$ links together a set of otherwise independent zero-sum games.

\begin{defin} The {\bf Dual Adversarial Data Augmentation} (ADA-D) game is:
\label{def:dual}
\begin{align}
& \min_{\theta} 
\mathbb{E}_{{\bf x},y^* \sim \mathcal{D}}
\Big[
\min_{f} \max_{P}
\mathbb{E}_{
 \arraycolsep=0pt\def\arraystretch{1.0}
 \tiny\begin{array}{l} 
 y'\sim f,\\ 
 y \sim P
\end{array}}
\Big[
\ell(y',y) \label{eq:dual} \\ & 
\qquad\qquad\qquad +
\theta^{\top} \{\phi(y,\vec{x})-\phi(y^*,\vec{x})\}
\Big]. \notag
\end{align}
\end{defin}
We make two important observations based on this dual optimization perspective.  
First, since $P$ is adversarially chosen, there is no need to restrict or parameterize $f$ to avoid overfitting to $P$ as is typically done in supervised learning.
Instead, the feature potential based on $\theta$ is learned to provide constraints on the adversary that make prediction easier.  Further, since $P$ is chosen \emph{after} $f$ in Eq. \eqref{eq:dual}, the predictor is incentivized to randomize.

\section{Adversarial Object Localization}

Up to this point, we have described our proposed adversarial data augmentation learning approach in general terms, as it can apply to many structured output tasks. Now, we shift our focus to the concrete problem of object detection. This explicit focus will help us to describe our approach in concrete terms.

\vspace{1mm}
\noindent \textbf{Label Space.} Each structured output label $y$ is represented by the four coordinates of a bounding box. The domain of a label is denoted $\mathcal{Y}$. The set of all possible bounding boxes $\mathcal{Y}$ is very large for an image of modest size and therefore it is rarely practical to evaluate all possible bounding boxes. This means that the sums over labels used in the formulation above (Section \ref{sec:formulation}) are not tractable and that some form of distribution approximation is needed. To discretize the label space $\mathcal{Y}$, we use a bounding box proposal algorithm, Edgebox \cite{edge-boxes-locating-object-proposals-from-edges}, to generate a set of $k$ bounding boxes to define the label space $\mathcal{Y}$.

\vspace{1mm}
\noindent \textbf{Feature Statistics.} To represent the feature statistics $\phi(y,\vec{x})$ of a bounding box $y$ over an image $\vec{x}$, we use the FC7 features of the VGG16 \cite{simonyan2014very} network. Concretely, it is a 4096 dimensional vector over a sub-image defined by the bounding box $y$. The feature statistic constraint $\{\phi(y',\vec{x})-\phi(y^*,\vec{x})\}$ described in the ADA-D definition represents the difference between the FC7 features of an arbitrary bounding box $y'$ and the FC7 features of the ground truth bounding box $y^*$. Also known as the \textit{perceptual loss}, this quantity ensures that the adversarial bounding box label remains perceptually similar to the ground truth bounding box label.

\vspace{1mm}
\noindent \textbf{Loss Function.} The loss function used for object localization is based on the classical intersection over union (IoU) score, $IoU(y,y') = {\text{area}(y \cap y')}/{\text{area}(y \cup y')}$. Here, $y$ and $y'$ are two bounding boxes. In this work, we focus on losses defined in terms of the amount of non-overlap, $\ell(y, y') = 1 - \text{IoU}(y,y')$,
which equals to one when $y$ and $y'$ are disjoint, zero when they are identical, and smoothly transitions in between those extremes. Another loss function we use is the overlap loss with a threshold:
\begin{align}
\ell_{t\alpha}(y,y') = \begin{cases}
1  &\text{IoU}(y,y') < \alpha\\
0  & \text{IoU}(y,y') \geq \alpha,
\end{cases} \label{eq:threshold}
\end{align}
which assigns binary loss to bounding boxes depending on the overlap threshold.

%==================================================================+%

\subsection{Game Matrix}

\begin{figure}[tb]
\centering
\includegraphics[width= 0.95\columnwidth]{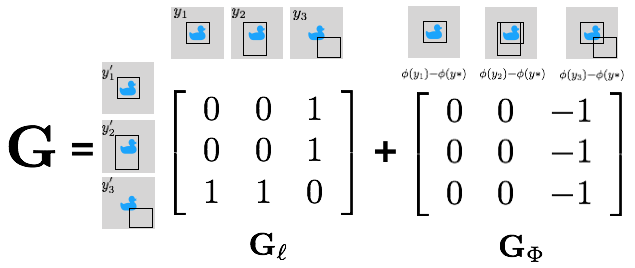}
\caption{Example Game Matrix for a duck image with three bounding boxes. Each black bounding box is a potential label for the same duck image.}
\label{fig:G}
\vspace{-3mm}
\end{figure}

As noted in Eq. \eqref{eq:matrix}, the expected loss of the adversarial game can be written in matrix form, $\mathbf{f}^{\top}\mathbf{G}\mathbf{p}$. The game (or payoff) matrix $\mathbf{G}$ for ADA-D can be constructed from Eq. \eqref{eq:dual} as an $|\mathcal{Y}|\times|\mathcal{Y}|$ matrix, where each element is defined as:
\begin{align}
    \mathbf{G}(y',y) 
    = \ell(y',y) + 
    \theta^{\top} 
    \{ 
      \phi(y,\vec{x}) - \phi(y^*,\vec{x}) 
    \},
\end{align}
where is the first term $\ell(\cdot,\cdot)$ is the IoU based loss and the second term is the weighted difference between FC7 features of the annotation distribution label $y$ and the ground truth $y^*$ label. To better understand the structure of the game matrix, we can decompose it as $\mat{G} = \mat{G}_{\ell} + \mat{G}_{\Phi}$. The elements of each matrix are illustrated for a toy example in Figure \ref{fig:G}. The first matrix $\mat{G}_{\ell}$ contains the pairwise loss between the label of the predictor and the label of the adversary, $\ell(y',y)$. The second matrix $\mat{G}_{\Phi}$ contains the difference in feature statistics between the adversarial label and the ground truth label, $\theta^{\top}\{\phi(y,\vec{x}) - \phi(y^*,\vec{x})\}$. Since the constraint matrix $\mat{G}_{\Phi}$ does not depend on the predictor label $y'$, each row is identical. The elements of the last column are all $-1$ in this example because the feature statistic of the bounding box $y_3$ are very different from the ground truth bounding box $y^*$, whereas the first two columns are zero because the content of their bounding boxes are similar.

\subsection{Nash Equilibria}

The solution of the game, the Nash equilibrium pair $({\bf f}, {\bf p})$, is defined as the optimal strategy for each player such that:
\begin{align}
\max_{{\bf p}'} {\bf f}^\top \mat{G} {\bf p}'
\leq {\bf f}^\top \mat{G} {\bf p} \leq
\min_{{\bf f}'} {{\bf f}'}^\top \mat{G} {\bf p}. \label{eq:nash}
\end{align}
For the example in Figure \ref{fig:G}, the lower potential for the third bounding box in $\mat{G}_\Phi$ offsets the larger loss that might be produced in $\mat{G}_\ell$ by having $p(y_3) > 0$.  
Due to the symmetries in $\mat{G}$, the equilibrium solution is then simply: $P(y_1|{\bf x})=P(y_2|{\bf x}) = f(y_1'|{\bf x})=f(y_2'|{\bf x})=0.5$ and $P(y_3|{\bf x})=f(y_3|{\bf x}) = 0$.

In general, this equilibrium solution pair can be obtained efficiently using a pair of linear programs:
\begin{align}
&\min_{v, {\bf f} \geq {\bf 0}} v \text{ such that: } {\bf f}^{\top} \mat{G} \leq v {\bf 1} \text{ and } {\bf f}^{\top} {\bf 1} = 1; \text{ and}
\label{eq:lp} \\
&\max_{v, {\bf p} \geq 0} v \text{ such that: } \mat{G} {\bf p} \geq v {\bf 1}^{\top} \text{ and } {\bf p}^{\top} {\bf 1} = 1, \notag
\end{align}
where 
$v$ is the value of the game (\emph{i.e.}, the expected loss).
This first linear program finds ${\bf f}$ that produces the maximum value against the worst choice of ${\bf p}'$ using the left-hand side of Eq. \eqref{eq:nash} via constraints for each deterministic choice of ${\bf p}'$ (i.e., the ${\bf 1}$ vector).  The second linear program is constructed in a likewise manner to obtain ${\bf p}$.

\subsection{Constraint Generation for Large Games}

In practice, forming and solving a zero-sum adversarial game over a very large set of labels (\emph{e.g.}, the set of all possible bounding boxes in an image) for each image is computationally expensive. To obtain the same result more efficiently, we employ a constraint-generation method \cite{mcmahan2003planning,wang2015adversarial} to solve ADA-D without explicitly constructing the entire payoff matrix $\mat{G}$. 
It is based on the key insight that the equilibrium distributions, ${\bf f}$ and ${\bf p}$, both assign zero probabilities to many bounding boxes and those bounding boxes can then be effectively removed from the game matrix without changing the solution.  
The basic strategy of constraint-generation is to use a set of the most violated constraints to grow a game matrix that supports the equilibrium distribution, but is much smaller than the full game matrix. %to optimize the objective function.

The approach works by iteratively obtaining a Nash equilibrium for a game defined over a subset of the possible labels (not all of them), finding a player's best response strategy (either the predictor or the annotation distribution) to that equilibrium distribution. Then the best response to the set of opponent strategies defining the game is added as a new strategy. When additional best responses no longer improve either player's game value, the subgame equilibrium is guaranteed to be an equilibrium to the larger game \cite{mcmahan2003planning}.

\setlength{\textfloatsep}{1mm}
\begin{algorithm}[tb]
\small
\caption{ ADA Equilibrium Computation}
\label{alg:doracle}
\begin{algorithmic}[1]
\REQUIRE{Image $\vec{x}$; Parameters $\theta$; Ground Truth $y^*$}
\ENSURE{Nash equilibrium,
$({\bf f}%(\hat{\bf y}|{\bf x})
,{\bf p}%(\check{\bf y}|{\bf x})
)$}
            \STATE $\mathcal{Y} \leftarrow \text{EdgeBox}(\vec{x})$
            \STATE $\Phi= \text{CNN}(\mathcal{Y},\vec{x})$
            %\State $\Phi_{groundTruth}= VggNet.Layer38(groundTruth)$
            \STATE $\boldsymbol{\psi} \leftarrow \theta^{\top} (\Phi -\Phi(y^*)) $
            %\State $\boldsymbol{\psi}_{groundtruth} \leftarrow \theta . \Phi_{groundtruth} $
            \STATE ${\mathcal{S}_p}\leftarrow {\mathcal{S}_f} \leftarrow  \argmax_{y} {\psi}(y)$ 
            \REPEAT{
                    \STATE $({\bf f},{\bf p},v_p) \leftarrow \texttt{solveGame}(%\hat{S},\check{S},
                    \psi({\mathcal{S}_p}),\text{loss}({\mathcal{S}_f},{\mathcal{S}_p}))$ 
                    \STATE $({y}_{\text{new}}, v_{\text{max}}) \leftarrow\! \max_{y}\mathbb{E}_{{y'} \sim f}[\text{loss}(y,y')\!+{\psi(y)}]$
                         \IF {$ (v_p \neq v_{\text{max}})$}
                                  \STATE ${\mathcal{S}_p} \leftarrow {\mathcal{S}_p} \cup y_{\text{new}}$
            \ENDIF
                    \STATE $({\bf f},{\bf p},{v}_f) \leftarrow \texttt{solveGame}(%\hat{S},\check{S},
                    \psi(\mathcal{S}_p),\text{loss}(\mathcal{S}_f,\mathcal{S}_p))$ 
                    \STATE $ ({y'}_{\text{new}}, v_{\text{min}}) \leftarrow \min_{\hat{y}} \mathbb{E}_{y \sim {p}}[ \text{loss}({y},{y'})]$
                        \IF {$ (v_f \neq v_{\text{min}})$}
                            \STATE ${\mathcal{S}}_f \leftarrow {\mathcal{S}}_f \cup {y'}_{\text{new}} $
                            \ENDIF
            }
        \UNTIL { ${v}_p = v_{\text{max}} = {v}_f = v_{\text{min}}$}
\RETURN 
$({\bf f}%(\hat{y}|{\bf x})
,{\bf p}%(\check{y}|x)
)$
\end{algorithmic}
\end{algorithm}

\subsection{Algorithm Details}

Algorithm 1 details the ADA equilibrium computation structure. The pre-processing step, extracting the image box proposals (EdgeBox) and their CNN features, is addressed in Lines 1-4. The CNN features are extracted from the last convolutional layer of the deep networks in respect with the related experiments.
$S_p$ and $S_f$ stand for strategies(chosen box proposals) of p and f game players.  The algorithm presents the main structure in lines 5-16. \texttt{solveGame} is the process of obtaining a Nash equilibrium using linear programming. Gurobi LP solver \cite{gurobi} is used to solve the linear programming. The constraint generation is presented with $\max$ and $\min$ operations in lines 7,12.
After reaching the loop termination condition (line 16), the $\mathbf{f}$ and $\mathbf{p}$ distributions are returned.

Model parameters $\theta$ are obtained using convex optimization (gradient-based methods \cite{duchi2011adaptive}) that reach convergence exactly when the data augmentations match the features of the training data:  $\mathbb{E}_{
\arraycolsep=0pt\def\arraystretch{1.0}
\tiny\begin{array}{l} 
\vec{x} \sim \mathcal{D},\\
y |\vec{x} \sim P
\end{array}}\left[\phi(y, \vec{x})\right] 
=
\mathbb{E}_{y,x \sim \mathcal{D}}
\left[\phi(y, \vec{x})\right]$.
The gradient is simply the difference between these two expectations.

At testing time, we employ a similar inference procedure.  It likewise uses constraint generation---in the form of best responses to equilibria using sets of strategies---to find a set of strategies supporting the equilibrium of the full game. A final prediction is produced by taking the most probable predictor strategy under the equilibrium distribution, $\hat{y} = \argmax_y P(y|\vec{x})$, as the predicted bounding box.

\section{Experiments}

In the following experiments, our aim is to show that our proposed adversarial optimization for data augmentation will provide meaningful improvements in test time prediction performance. We first compare our adversarial data augmentation (ADA) objective function against baselines models with no data augmentation on the task of localization in Section \ref{sec:noaugment}. Second, we compare our approach to baseline models with varying levels of data augmentation in Section \ref{sec:augment}. Third, we evaluate our approach on the task of detection (joint recognition and localization) in Section \ref{sec:detection}. Finally, we use our adversarial optimization over various deep features to show consistent improvements across networks in Section \ref{sec:arch}.

\vspace{1mm}
\noindent\textbf{Baselines}. To analyze the performance of our adversarial objective function (ADA) for object detection, we benchmark it against two classical objective functions.

\noindent\textbf{(1) SSVM:} The structured output support vector machine (SSVM)\cite{vedaldi11svm-struct-matlab} is a large margin classifier with a variable margin depending on a structured loss function $\ell$. The objective function is defined as:
\begin{align}
\hat{\theta} &= \argmin_{\theta} ~~~  \lambda || \theta ||_2 + \sum_n \xi_n\\
\textrm{s.t.} ~~~ 
& ~\theta^{\top} (\phi (y^*_n,\vec{x}_n) - \phi (y,\vec{x}_n) ) \geq \ell (y^*_n,y) - \xi_n ~~~ \forall ~ y \nonumber
\end{align}
where $\theta$ is the weight vector, $\phi$ is the feature function (image feature statistic), $\ell$ is the loss function and $\xi$ is the slack variable. To solve the SSVM objective function, we employ an iterative constraint generation strategy to accelerate the learning process by adding a few constraints per iteration (instead of the entire constraint set defined by each label $y\in\mathcal{Y}$). At test time, we generate a set of bounding box proposals (EdgeBox) and take the bounding box with the highest potential using the learned weight vector, $\hat{y} = \argmax_{y} \theta^{\top}\phi(y,\vec{x})$ to identify the predicted structured output.
 
\noindent\textbf{(2) Softmax:} The soft maximum (logistic regression) objective function is a probabilistic predictor. For the softmax objective function, we estimate a distribution over all proposed bounding boxes $y$ that maximizes the conditional likelihood of proposed bounding boxes with an IoU above a given threshold.
\begin{align}
    \hat{\theta}
    &=\argmax_{\theta} \prod_n P(y_n|\vec{x}_n;\theta),\\
    &=\argmax_{\theta} \prod_n 
    \frac
    {e^{\theta^{\top}\phi(y_n,\vec{x}_n)}}
    {\sum_{y} e^{\theta^{\top}\phi(y,\vec{x}_n)} }
\end{align}
where $\theta$ is the weight vector, $\phi$ is the potential function (FC7 feature) and $\ell$ is the loss function. At test time, we compute the Bayesian optimal decision to identify the most likely bounding box from a set of proposed bounding boxes according to:
\begin{align}
\hat{y} = \argmin_{y} \sum_{y' \in \mathcal{Y}} P(y'|{\bf x};\theta) \ell(y, y'), \label{eq:bayesopt}
\end{align}
where $ P(y|{\bf x};\theta)$ is the learned conditional distribution parameterized by $\theta$ and $\ell$ is the loss function.

\begin{table}[tb]
\begin{center}\small
\caption{No augmentation baseline comparison (IoU$>$0.5)}
\label{tab:noaugment50}
\vspace{-2mm}
\setlength\tabcolsep{2pt}
\scalebox{0.73}{
\begin{tabular}{| l | r | r | r | r | r | r | r | r | r | r || r | }
\hline
\multirow{2}{*}{\;{\bf Model}} &  \multicolumn{10}{c||}{{\bf ImageNet Object Categories}} & \\
 \cline{2-12} & Plane & Bird & Bus & Car & Cat & Cow & Dog & Hors & Moni & Sofa & mAP\\\hline
ADA+VGG (Ours) & { \bf 92.0} & { \bf 93.5} & {\bf 92.0} & {\bf 100.0} & {\bf 89.1} & {\bf 100.0} & {\bf 93.0} & {\bf 96.4} & {\bf 96.0} & {\bf 90.0} & {\bf 94.2}\\
Softmax+VGG&	84.0 &	86.5 &	84.0 & 87.0 & 70.9 & 77.5 & 62.0 & 72.7 & 72.0 & 80.0 & 77.7
\\
SSVM+VGG & 90.0 & 82.5 & 82.0 & 82.0 & 40.0 & 87.5 & 72.0 & 72.7 & 90.0 & 78.0 & 77.7
\\\hline
\end{tabular}
}
\end{center}
\vspace{-3mm}
\end{table}

\begin{table}[tb]
\begin{center}
\small
\caption{ No augmentation baseline comparison (IoU$>$0.7)}
\vspace{-2mm}
\label{tab:noaugment70}
\setlength\tabcolsep{2pt}
\scalebox{0.75}{
\begin{tabular}{| l | r | r | r | r | r | r | r | r | r | r || r | }
\hline
\multirow{2}{*}{\;{\bf Model}} &  \multicolumn{10}{c||}{{\bf ImageNet Object Categories}} & \\
 \cline{2-12} 
 & Plane & Bird & Bus & Car & Cat & Cow & Dog & Hors & Moni & Sofa & mAP\\\hline
%VGG + ADA$_{50}$ & {\bf 52.0} & 45.0 & {\bf 44.0} & {\bf 72.0} & 16.4 & {\bf 67.5} & {\bf 32.0} & 29.1 & {\bf 52.0} & 38.0 &{\bf 44.8}\\
%VGG + Softmax$_{50}$ &  38.0 & 33.0 & 20.0 & 45.0 & 12.7 & 12.5 & 14.0 & 09.1 & 14.0 & 22.0 & 22.0\\
%VGG + SSVM$_{50}$ & 46.0 & {\bf 48.0} & 30.0 & 45.0 & {\bf 20.0} & 40.0 & 25.0 & {\bf 30.9} & 42.0 & {\bf 42.0} & 36.9\\
%\hline
ADA+VGG (Ours) & {\bf 58.0} & {\bf 61.5} & {\bf 64.0} & {\bf 91.0} & {\bf 30.9} & {\bf 77.4} & {\bf 58.0} & {\bf 58.2} & {\bf 61.8} & {\bf 61.9} & {\bf 62.3}
\\
Softmax+VGG  &	47.6 &	45.7 &	40.0 &	62.8 &	20.0 &	42.5 &	25.1 &	25.4 &	31.4 &	44.2 & 38.5
\\
SSVM+VGG  & 51.8 & 55.5 & 44.0 & 61.7 & 21.8 & 54.7 & 31.6 & 43.6 & 56.0 & 57.3 & 47.8\\
%ADA & 72.6 & 72.9 & 72.1 & 80.8 & 66.2 & 78.3 & 71.1 & 69.1 & 73.6 & 74.1 & ...\\
%Softmax & 67.5 & 68.8 & 66.5 & 71.9 & 60.6 & 64.7 & 61.5 & 63.3 & 64.8 & 68.2 & ...\\
%SSVM & 69.6 & 70.9 & 67.3 & 72.9 & 57.2 & 70.6 & 63.7 & 67.2 & 70.1 & 69.6 & ...
\hline
\end{tabular}
}
\end{center}
\vspace{-3mm}
\end{table}

\subsection{Baseline Comparisons with No Augmentation}\label{sec:noaugment}

We begin with the simplest evaluation, where we compare our proposed adversarial data augmentation approach with two baseline method that use only the ground truth annotation, without augmenting the training data, to learn a predictor. We compare our method \textbf{ADA+VGG} against \textbf{SSVM+VGG} and \textbf{Softmax+VGG}. The suffix \textbf{+VGG} for each objective function specifies the deep network from which the features are used, in this case VGG16 \cite{simonyan2014very}. We compute the mean Average Precision (mAP) score for several classes in ImageNet dataset for each of the competing methods. We train a bounding box predictor for each object category, and consider an object to be correctly detected when the IoU is greater than a threshold. We emphasize here that we are decoupling the recognition task from the localization task by learning class specific bounding box regressor and testing only on images that contain the target class. Later experiments will evaluate on both recognition and localization. For this experiment, we give results for two thresholds, 50\% IoU and 70\% IoU for each object category. Since our method explicitly augments the dataset as part of the optimization process whereas the two baselines have no data augmentation, we expect our approach will outperform the two baselines.

The test time localization accuracy at 50\% IoU on object 10 classes from ImageNet are given in Table \ref{tab:noaugment50}. As expected, we observed significant test time improvement in bounding box regression accuracy for every object category. On average, our proposed adversarial data augmentation approach improved mean average precision by 17\% percentage points.

We repeated the same experiment for a more strict loss, a {0.7} thresholded IOU loss function. The mean average precision of the predicted bounding boxes are given in Table \ref{tab:noaugment70}. Since we are evaluating performance with a more strict loss function, the absolute mAP values decrease as expected. However, notice that our proposed approach still obtains a significant improvement over the baseline algorithms improving mAP by 15\% percentage points over the strongest baseline \textbf{SSVM+VGG}.

\subsection{Baseline Comparisons with Augmentation}\label{sec:augment}

We now compare the performance of our approach to the strongest baseline model, \textbf{SSVM+VGG}, trained with different levels of data augmentation. As mention earlier, data augmentation such as random translations of bounding boxes, is a common heuristic used to help supervised learning methods avoid overfitting. We prepare five levels of data augmentation to train the \textbf{SSVM+VGG} baseline. Instead of using random translations within a range of the ground truth bounding box annotation, which can contain many similar bounding boxes, we use the EdgeBox proposal network to generate a diverse set of bounding boxes. We keep the top 250 EdgeBox proposals with the highest scores and filter them according to 5 thresholds with respect to the original ground truth bounding box: (1) IoU$>$50\%; (2) IoU$>$60\%; (3) IoU$>$70\%; (4) IoU$>$75\%; and (5) IoU$>$80\%. We denote the experiment using the subscript \texttt{t50} to represent a model trained on a collection of bounding boxes with IoU$>$50\%.  We consider the bounding boxes that pass the threshold test, as new `ground truth' and use them as the training set. We note here again that our proposed method automatically selects (gives weights to) the bounding box proposals during the learning process and does not require a separate augmentation step.

The results of this experiment are shown in Table \ref{tab:perturb}.  We note that augmenting SSVM in this manner improves test performance compared to the model without data augmentation (Table \ref{tab:noaugment70}). We also observe that the performance gain is maximized around the 70\% overlap threshold.  However, we find that with the exception of the \emph{Bird} class, the performance gains of data augmentation do not reach the performance of our ADA approach. Our proposed approach still outperforms the model with the best level data augmentation by $6.4\%$ percentage points.

\begin{table}[tb]
\begin{center}\small
\caption{Effect of Data Augmentation (IoU $>$ 70\%)}\vspace{1mm}
\label{tab:perturb}
\setlength\tabcolsep{2pt}
\scalebox{0.75}{
\begin{tabular}{| l | r | r | r | r | r | r | r | r | r | r || r | }
\hline
\multirow{2}{*}{\;{\bf Augmentation}} &  \multicolumn{10}{c||}{{\bf AlexNet Object Category}} & \\
 \cline{2-12}
 & Plane & Bird & Bus & Car & Cat & Cow & Dog & Hors & Moni & Sofa & Avg\\\hline
SSVM$_{t50}$+VGG & 53.8 & 57.9 & 49.7 & 64.0 & 22.6 & 59.9 & 37.5 & 45.5 & 56.7 & 57.8 & 50.5 \\
SSVM$_{t60}$+VGG & 54.7 & 58.9 & 52.7 & 67.7 & 23.7 & 64.9 & 42.0 & 48.6 & 57.3 & 58.4 & 52.9 \\
SSVM$_{t70}$+VGG & {\bf 56.4} & {\bf 61.6} & {\bf 56.8} & {\bf 70.8} & {\bf 25.4} & {\bf 67.3} & {\bf 49.1} & {\bf 51.9} & {\bf 58.6} & {\bf 58.8} & {\bf 55.7} \\
SSVM$_{t75}$+VGG & 52.6 & 61.0 & 51.7 & 64.4 & 20.2 & 61.2 & 42.6 & 44.0 & 57.3 & 56.0 & 51.1 \\
SSVM$_{t80}$+VGG & 49.8 & 52.0 & 44.9 & 60.3 & 20.2 & 55.8 & 33.1 & 41.4 & 55.8 & 52.7 & 46.6 \\
\hline
ADA+VGG (Ours) & {\bf 58.0} & {\bf 61.5} & {\bf 64.0} & {\bf 91.0} & {\bf 30.9} & {\bf 77.4} & {\bf 58.0} & {\bf 58.2} & {\bf 61.8} & {\bf 61.9} & {\bf 62.3}
\\
\hline
\end{tabular}}
\end{center}
\vspace{-3mm}
\end{table}

We also performed a second data augmentation experiments by only varying the number of bounding boxes with top the EdgeBox scores (instead of using IoU) for every image label to understand how the amount of augmented data affects test time performance. At test time, we use the 50\% overlap criteria for successful localization. Table \ref{tab:ktop} shows the mAP performance over same 10 object categories in ImageNet as a function of the number of augmented data annotations per image. The augmented data annotations are selected from a rank list of EdgeBox proposals from each image. The performance of the \textbf{SSVM+VGG} tops out at 8 augmented data annotations and is still 12\% points below our propose approach (94.2\% mAP).

\begin{table}[tb]
    \centering
    \caption{Effect of Number of Augmented Data Annotations. ADA outperforms best configuration SSVM+VGG baseline by 12\%.}\vspace{1mm}
    \label{tab:ktop}
    \setlength\tabcolsep{2pt}
    \scalebox{0.85}{
    \begin{tabular}{|c|c|c|c | c|c|c|c||c|}\hline
         SSVM+VGG           & k=1 & k=2   & k=4  & k=6  & k=8  & k=10  & k=12 & ADA+VGG\\\hline
            mAP & 77.6    & 79.7  & 81.4 & \bf 83.8 & 83.7 & 79.8  & 75.3  & \bf 94.2\\\hline
    \end{tabular}
    }
\end{table}

\begin{figure*}
\begin{center}
{\bf Adversarial Boxes (proposed)}\\
\includegraphics[width=0.15\textwidth]{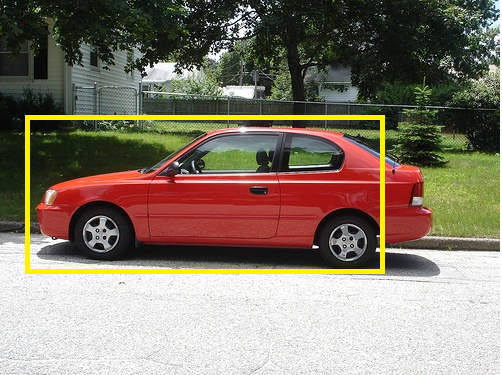}
\includegraphics[width=0.15\textwidth]{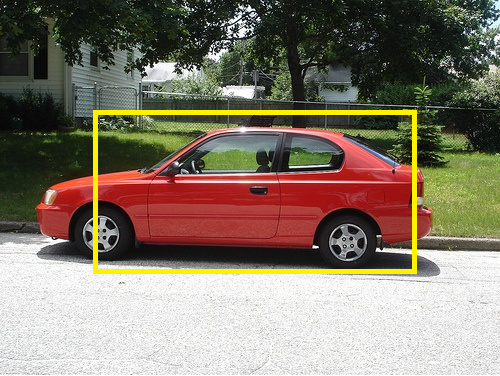}
\includegraphics[width=0.15\textwidth]{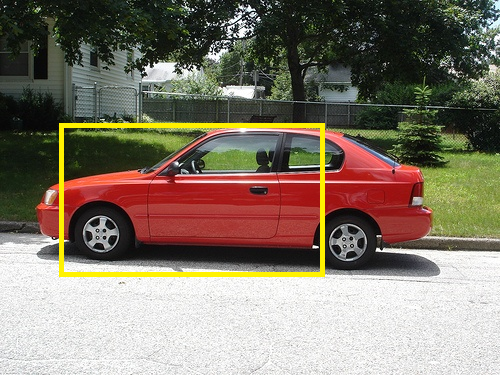}
\includegraphics[width=0.15\textwidth]{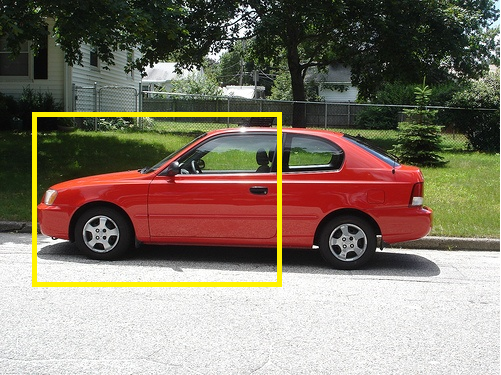}
\includegraphics[width=0.15\textwidth]{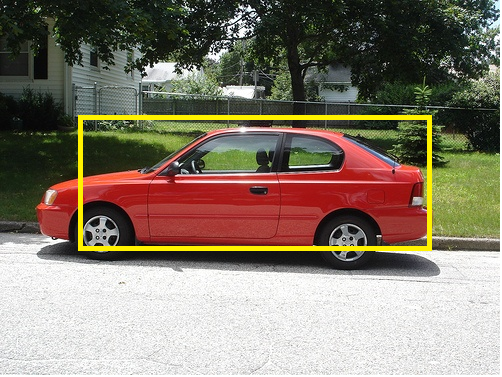}\\
\vspace{3mm}
{\bf Top-K EdgeBoxes by score}\\
\includegraphics[width=0.15\textwidth]{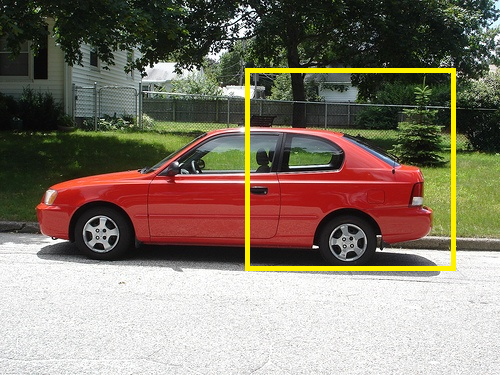}
\includegraphics[width=0.15\textwidth]{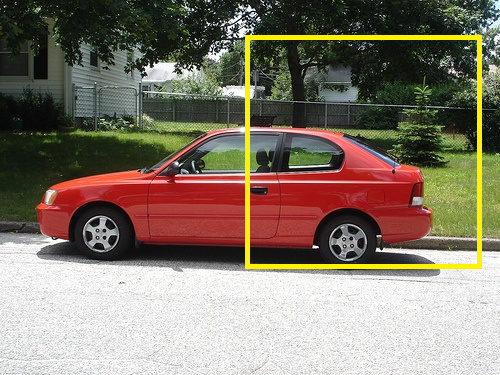}
\includegraphics[width=0.15\textwidth]{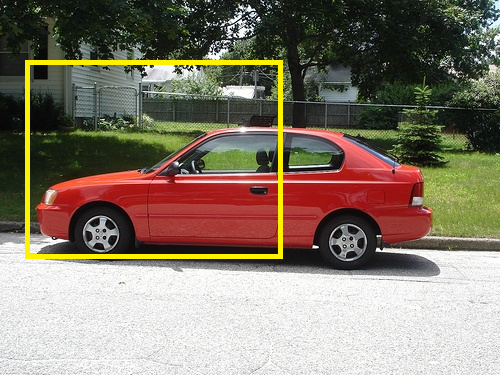}
\includegraphics[width=0.15\textwidth]{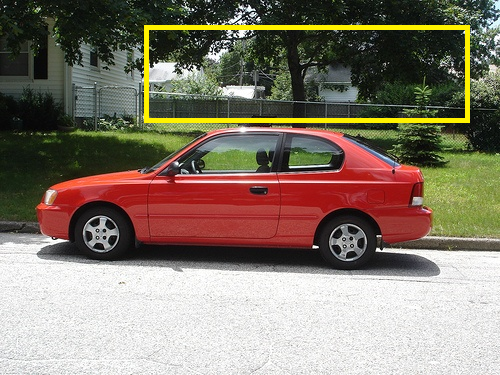}
\includegraphics[width=0.15\textwidth]{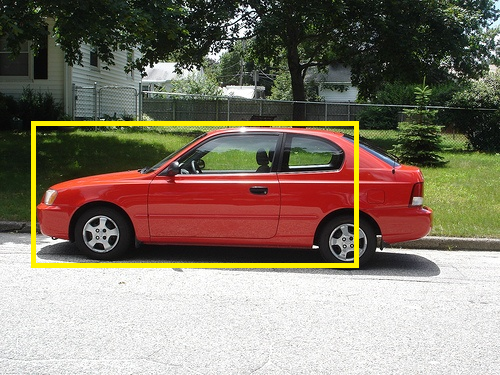}
\\
\vspace{3mm}
{\bf Top-K EdgeBoxes by IoU}\\
\includegraphics[width=0.15\textwidth]{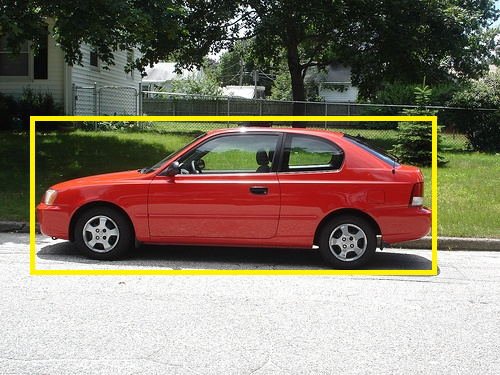}
\includegraphics[width=0.15\textwidth]{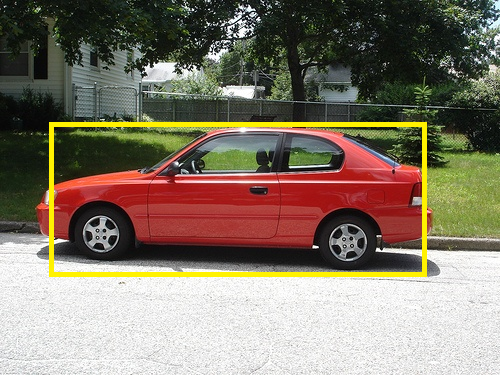}
\includegraphics[width=0.15\textwidth]{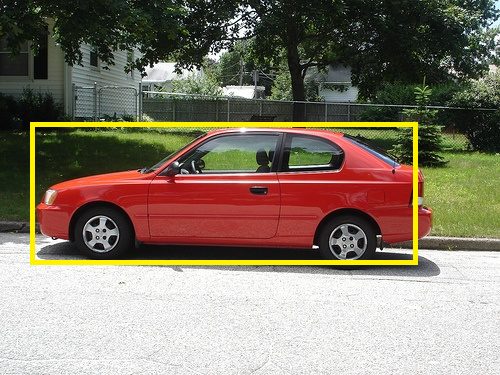}
\includegraphics[width=0.15\textwidth]{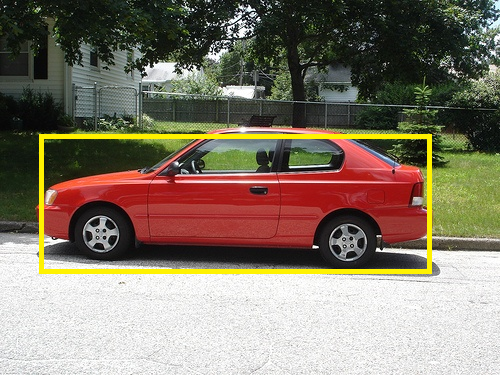}
\includegraphics[width=0.15\textwidth]{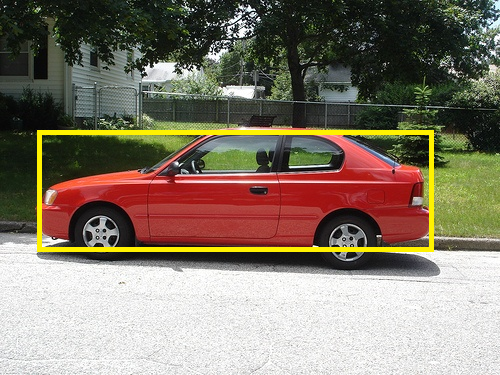}
\caption{The top five augmented bounding boxes (left to right) under the different augmentation strategies. Our adversarial augmentation approach generates much more diverse bounding boxes to augment the training data than choosing solely based on IoU yet the boxes are at the same time still relevant to the target object unlike general EdgeBoxes.}
\label{fig:bboxes}
\end{center}
\end{figure*}

\begin{table}[tb]
\begin{center}
\caption{Detection Performance Comparison (IoU $>70\%$).}\vspace{1mm}
\small
\label{tab:task4a}
\setlength\tabcolsep{2pt}
\scalebox{0.75}{
\begin{tabular}{| l | r | r | r | r | r | r | r | r | r | r || r | }
\hline
\multirow{2}{*}{\;{\bf Model}} &  \multicolumn{10}{c||}{{\bf Image Net Object Category}} & \\
 \cline{2-12}
 & Plane & Bird & Bus & Car & Cat & Cow & Dog & Hors & Moni & Sofa & Avg\\\hline
ADA+VGG (Ours) & {\bf 46.0} & {\bf 55.5} & {\bf 60.0} & {\bf 86.0} & {\bf 25.4} & {\bf 70.0} & {\bf 47.0} & {\bf 52.7} & {\bf 60.0} & {\bf 48.0} & {\bf 55.1} \\
SSVM+VGG & 42.0 & 46.0 & 38.0 & 53.0 & 16.4 & 52.5 & 25.0 & 36.4 & 42.0 & 42.0 & 39.3 \\
Softmax+VGG & 40.0 & 42.5 & 42.0 & 55.0 & 16.4 & 32.5 & 16.0 & 29.1 & 22.0 & 34.0 & 33.0 \\
\hline
\end{tabular}}
\end{center}
\vspace{-3mm}
\end{table}

\begin{table*}[t]
\begin{center}
\caption{ADA Generalization Across Deep Architectures. VOC2007 mAP for IoU$>$0.5.}
\label{tab:task2a}
\setlength\tabcolsep{2pt}
\scalebox{0.8}{
\begin{tabular}{| l | r | r | r | r | r | r | r | r | r | r | r | r | r | r | r | r | r | r | r | r || r |}
\hline
\multirow{2}{*}{\;{\bf Model}} &  \multicolumn{20}{c||}{{\bf VOC 2007 Object Category}} & \\
 \cline{2-22} & Aero & Bike & Bird & Boat & Bott & Bus & Car & Cat & Chair & Cow & DinT & Dog & Horse & mbike & person & Plant & Sheep & Sofa & Train & TV & mAP\\\hline
ADA+VGG16 & 68.5 & 71.5 & 67.8 & 63.3 & 48.6 & 76.5 & 78.8 & 80.9 & \bf 50.9 & 78.5 & 64.5 & 79.6 & 71.8 & 73.2 & 66.4 & 30.2 & 70.6 & 72.6 & 80.8 & 62.8 & 67.9\\\hline
SSVM+VGG16 & 73.6 & 76.4 & 63.7 & 46.1	&44.0&76.0&78.4&80.0&41.6& 74.2& 62.8 &79.8& 78.0& 72.5&64.3&35.0&67.2&67.2&70.8&\bf 71.4&66.1\\\hline
SVM+VGG16 \cite{girshick2014richTR}&73.4 &\bf 77.0 &63.4 &45.4 &44.6 &75.1 &78.1 &79.8 &40.5 &73.7 &62.2 &79.4 & \bf 78.1 &73.1 &64.2 &35.6 &66.8 &67.2 &70.4 &71.1 &66.0\\ \hline\hline
ADA+AlexNet fc7& 62.4 & 70.0 & 63.6 & 63.0 & 44.8 &  72.2 & 75.5  & 79.5 & 44.6 & 81.6 & 64.0 & 81.5& 70.2 & 68.5 & 71.4 & \bf 69.5& 65.0 & 71.2 & 81.4 &59.8 & 68.0\\\hline
SSVM+AlexNet fc7& 68.2 & 72.9 & 57.3 & 44.2 	&41.8&66.0&74.3&69.2&34.6&54.7&54.3&61.3&69.8&68.7&58.5&34.6&63.6&52.5&62.6 &63.5 &58.6\\\hline
SVM+AlexNet fc7 \cite{girshick2014richTR} &68.1 &72.8 &56.8 &43.0 &36.8 &66.3 &74.2 &67.6 &34.4 &63.5 &54.5 &61.2 &69.1 &68.6 &58.7 &33.4 &62.9 &51.1 &62.5 &64.8 &58.5 \\\hline\hline
ADA+ResNet101 &\textbf{76.4} & 74.8 &\textbf{72.4} & \textbf{64.0} & \bf 52.5 & \bf 84.0 & \bf 81.9 & \bf 86.0 & 48.5 & \bf 83.5 & \bf 64.8 & \bf 82.0 & 73.5 & \bf 77.0 & \bf 72.4 & 36.6 & \bf 74.4 & \bf 74.8 & \bf 81.4 & 65.6 & \bf 71.3 \\
 \hline
SSVM+ResNet101 & 68.0 & 70.2 & 69.3 & 54.3 & 46.5 & 76.2 & 78.8 &85.0&46.8& 80.2& 63.2& 78.1&69.5&71.4&61.8&36.8&68.1&  69.1&73.6&64.5& 66.5\\\hline
\end{tabular}}
\end{center}
\end{table*}

\subsection{Detection Performance Comparison}\label{sec:detection}

We now address the object detection task of jointly locating and recognizing the category an unknown object, to evaluate the performance of our approach on a harder task. In order for a model to obtain a correct result, the predictor must output the correct category label and also generate a bounding box that overlaps with the ground truth by at least 70\% IoU. For our baseline models, \textbf{SSVM+VGG} and \textbf{Softmax+VGG}, we use the best performing data augmentation scheme from Table \ref{tab:perturb} that includes EdgeBox proposals that have 70\% IoU threshold with the original ground truth annotation.

Table \ref{tab:task4a} shows the object detection performance when evaluated at the 70\% IoU threshold for correctness. We again find strong support for our adversarial approach to dealing with uncertainty. Specifically, we find that ADA$_{70}$ provides the best performance for all object classes. Though the relative performance advantage differs by object type, for classes like {\tt Dog}, the improvement over the other approaches is nearly double. On average, ADA provides a significant performance improvement of $15.8\%$ percentage points on this task over the strongest performing \textbf{SSVM+VGG} baseline.

\subsection{Generalization Across Deep Features}\label{sec:arch}

To understand how our proposed adversarial loss function, ADA, generalizes across various deep architectures, we compare mAP over 20 categories from the VOC2007 dataset for several combinations of deep architecture using our proposed adversarial loss function. In the pre-proccessing step of theses experiments, the box proposals of every image are extracted using EdgeBox. These box proposals are employed as input to the respective deep networks.
The CNN features are extracted from the last convolutional layer of the respective deep network and these box proposals and their CNN features are employed in training the respective classifier. We report the generalized version of ADA and SSVM with Alexnet\cite{krizhevsky2012imagenet}, VGG, and ResNet101\cite{he2016deep}.
For the VOC 2007 dataset, we use 5000 training images and 4952 testing images.

\section{Conclusions}

In this paper, we have developed a game-theoretic formulation for data augmentation that perturbs image annotations adversarially. This provides robustness in the learned predictor that is achieved by training from a number of augmentations that are adaptively selected to be difficult, while still approximating the ground truth annotation. We demonstrated the benefits for object localization and detection using experiments over ten different object classes for the ILSVRC2012 dataset and twenty different object classes for the VOC2007 dataset, showing significant improvements  for our approach under 50\% and 70\% thresholded IoU evaluation measures.

{\small
\bibliographystyle{ieee}
\bibliography{egbib}
}

\end{document}